\documentclass[twoside]{article}

\usepackage{lineno,hyperref}
\usepackage{multirow}
\usepackage{amsthm,amsmath}
\usepackage{amssymb}
\usepackage[ruled,linesnumbered, vlined]{algorithm2e}
\usepackage{subfigure}
\usepackage{amsfonts}
\usepackage{graphicx}
\usepackage{array}
\usepackage{booktabs}
\SetKwComment{commentSt}{! }{}

\usepackage{amsbsy}
\usepackage{textcomp}
\usepackage{marvosym}
\usepackage{picins}
\usepackage{caption}
\usepackage{threeparttable}
\usepackage{float}
\usepackage{lastpage}
\usepackage{lscape}

\usepackage{eurosym}
\usepackage{mathrsfs}
\usepackage{fancyhdr}
\usepackage{CJK}
\usepackage{multicol}
\usepackage{graphics}
\usepackage{indentfirst}
\usepackage{color}
\usepackage{bm}
\usepackage{upgreek}
\usepackage{booktabs}
\usepackage{warpcol}
\usepackage{epstopdf}

\def\footnoterule{\kern 1mm \hrule width 10cm \kern 2mm}

\allowdisplaybreaks
\catcode`@=11
\def\title#1{\vspace{3mm}\begin{flushleft}\vglue-.1cm\Large\bf\boldmath\protect\baselineskip=18pt plus.2pt minus.1pt #1
\end{flushleft}\vspace{1mm} }
\def\author#1{\begin{flushleft}\normalsize #1\end{flushleft}\vspace*{-4pt} \vspace{3mm}}

\def\jz#1#2{{$^{\footnotesize\textcircled{\tiny #1}}$\let\thefootnote\relax\footnotetext{\!\!$^{\footnotesize\textcircled{\tiny #1}}$#2}}}
\catcode`@=11
\def\section{\@startsection{section}{1}{\z@}%
 {-3ex \@plus -.3ex \@minus -.2ex}%
 {2.2ex \@plus.2ex}%
{\normalfont\normalsize\protect\baselineskip=14.5pt plus.2pt minus.2pt\bfseries}}
\def\subsection{\@startsection{subsection}{2}{\z@}%
 {-3ex\@plus -.2ex \@minus -.2ex}%
 {2ex \@plus.2ex}%
{\normalfont\normalsize\protect\baselineskip=12.5pt plus.2pt minus.2pt\bfseries}}
\def\subsubsection{\@startsection{subsubsection}{3}{\z@}%
 {-2.2ex\@plus -.21ex \@minus -.2ex}%
 {1.4ex \@plus.2ex}
{\normalfont\normalsize\protect\baselineskip=12pt plus.2pt minus.2pt\sl}}

\begin{document}
\begin{CJK*}{GBK}{song}
\thispagestyle{empty}
\vspace*{-13mm}
\vspace*{2mm}

\title{Knowledge Reasoning via Jointly Modeling Knowledge Graphs and Soft Rules}

\author{Yinyu Lan, Shizhu He, Kang Liu, Jun Zhao}
Institute of Automation, Chinese Academy of Sciences, Beijing 100190, China.\\

\let\thefootnote\relax\footnotetext{{}\\[-4mm]\indent\ Regular Paper}

\noindent {\small\bf Abstract} \quad  {\small {Knowledge graphs (KGs) play a crucial role in many applications, such as question answering, but incompleteness is an urgent issue for their broad application. Much research in knowledge graph completion (KGC) has been performed to resolve this issue. The methods of KGC can be classified into two major categories: rule-based reasoning and embedding-based reasoning. The former has high accuracy and good interpretability, but a major challenge is to obtain effective rules on large-scale KGs. The latter has good efficiency and scalability, but it relies heavily on data richness and cannot fully use domain knowledge in the form of logical rules. We propose a novel method that injects rules and learns representations iteratively to take full advantage of rules and embeddings. Specifically, we model the conclusions of rule groundings as 0-1 variables and use a rule confidence regularizer to remove the uncertainty of the conclusions. The proposed approach has the following advantages: 1) It combines the benefits of both rules and knowledge graph embeddings (KGEs) and achieves a good balance between efficiency and scalability. 2) It uses an iterative method to continuously improve KGEs and remove incorrect rule conclusions. Evaluations on two public datasets show that our method outperforms the current state-of-the-art methods, improving performance by 2.7\% and 4.3\% in mean reciprocal rank (MRR).}}

\vspace*{3mm}

\noindent{\small\bf Keywords} \quad {\small Distributed representation, Knowledge graph, Link prediction, Logical rule
}

\vspace*{4mm}

\end{CJK*}
\baselineskip=18pt plus.2pt minus.2pt
\parskip=0pt plus.2pt minus0.2pt
\begin{multicols}{2}

\section{Introduction}
\label{sec1}

A knowledge graph (KG) organizes knowledge as a set of interlinked triples, and a triple ({\it (head entity, relation, tail entity)}, simply represented as {\it (h, r, t)}) indicates the fact that two entities have a certain relation. Rich structured and formalized knowledge has become a valuable resource to support downstream tasks, for example, question answering~\cite{qa1, qa2} and recommender systems~\cite{rec1, rec2}.

Although KGs such as DBpedia~\cite{dbpedia}, Freebase~\cite{fb} and NELL~\cite{nell} contain large amounts of entities, relations, and triples, they are far from complete, which is an urgent issue for their broad application. To address this, the task of knowledge graph completion (KGC) has been proposed and has attracted growing attention; it utilizes knowledge reasoning techniques to perform automatic discovery of new facts based on existing facts in a KG \cite{zs1}.

At present, the methods of KGC can be classified into two major categories: 1) One type of method uses explicit reasoning rules; it obtains the reasoning rules through inductive learning and then deductively infers new facts. 2) Another method is based on representation learning instead of directly modeling rules, aiming to learn a distributed embedding for entities and relations and perform generalization in numerical space.

Rule-based reasoning is accurate and can provide interpretability for the inference results. Domain experts can handcraft these rules \cite{rmn} or  can mine them from the KG with an induction algorithm such as AMIE \cite{amie}. Traditional methods such as expert systems \cite{es,ites} use hard logical rules to make predictions. For example, as shown in Fig. \ref{fig1}, given the logical rule $ (x, born\_in, y) \wedge (y,city\_of,z)\Rightarrow (x, nationality, z) $ and the two facts that (Chicago, city\_of, USA) and (Mary, born\_in, Chicago), we can infer the fact (Mary, nationality, USA). A large number of new facts (conclusions) can be derived based on forward chaining inference. However, for large-scale KGs, sufficient and effective reasoning rules are difficult and expensive to obtain. Moreover, in many cases, the logical rules may be imperfect or even self-contradictory. Therefore, it is essential to model the uncertainty of (soft) logical rules effectively.

The methods of determining KGEs learn to embed entities and relations into a continuous low-dimensional space \cite{zs2,zs3}. These embeddings retain the semantic meaning of entities and relations, which can be used to predict missing triples. In addition, they can be effectively trained using stochastic gradient descent. However, this kind of method cannot fully use logical rules, which compactly encode domain knowledge and are helpful in various applications. Good embedding relies heavily on data richness, so these methods have difficulty learning useful representations for sparse entities \cite{sparsity,itere}.

In fact, both rule-based methods and embedding-based methods have advantages and disadvantages in the KGC task. Logical rules are accurate and interpretable, and embedding is flexible and computationally efficient. To achieve more precise knowledge completion, recently, there has also been research on combining the advantages of logical rules and KGEs. Mixed techniques can infer missing triples effectively by exploiting and modeling uncertain logical rules. Some existing methods have aimed to iteratively learn KGEs and rules \cite{itere}, and some other methods also utilize soft rules or groundings of rules to regularize the learning of KGEs \cite{sole,slre}.

\vspace{2mm}

\begin{center}
    \includegraphics[width=0.48\textwidth]{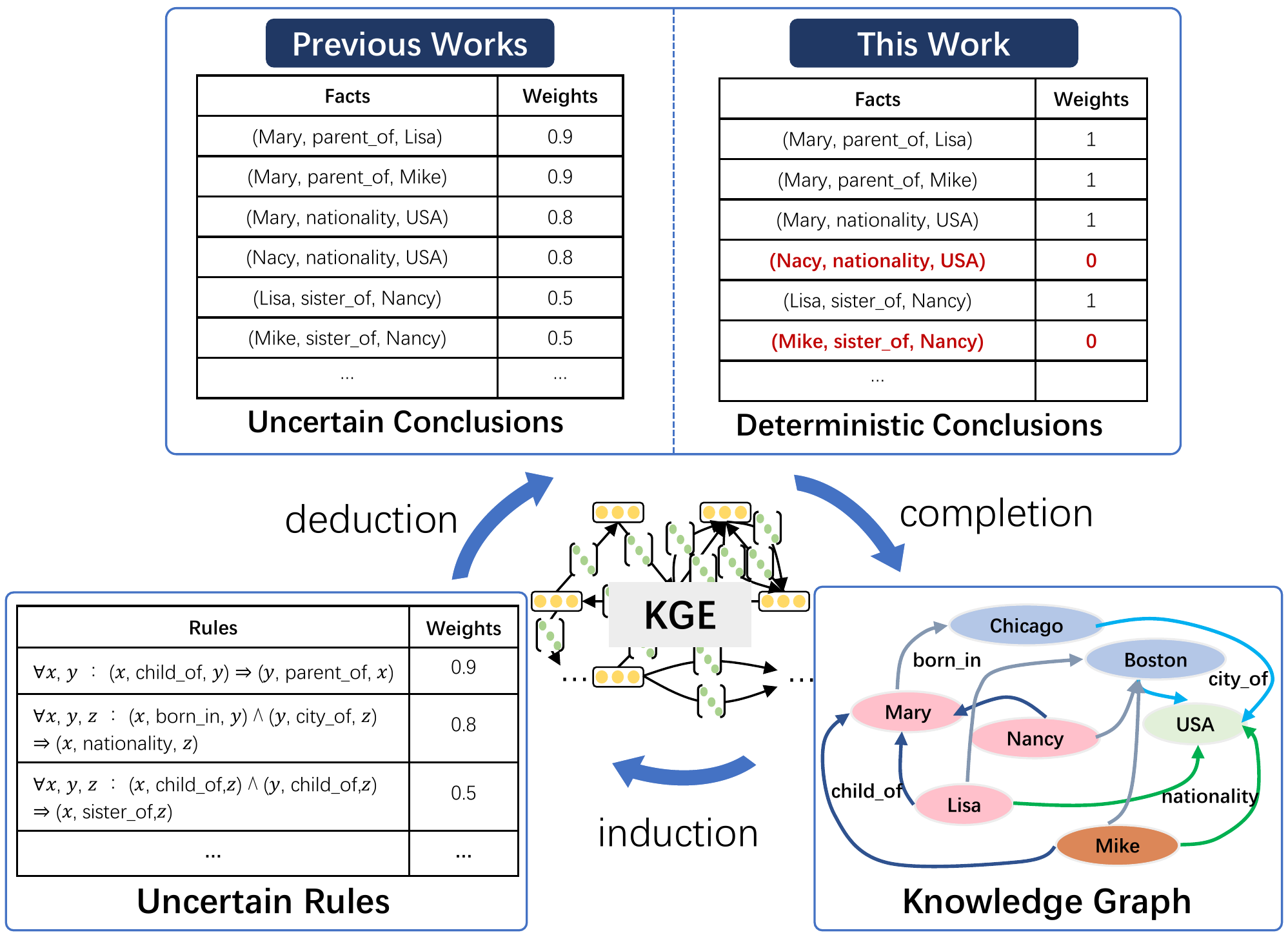}\\
    \vspace{2mm}
    \parbox[c]{8.3cm}{\footnotesize{Fig. 1.~} We propose a novel iterative knowledge reasoning framework by fusing logical rules into a KGE. Previous methods associate each conclusion with a weight derived from the corresponding rule. In contrast, our method can infer which conclusion is true via jointly modeling the deterministic KG and uncertain soft rules. }
    \label{fig1}
\end{center}

The integration of logical rules and knowledge graph embeddings can achieve more efficient and accurate knowledge completion. Current methods model uncertain rules and add soft labels to conclusions by t-norm-based fuzzy logic \cite{tnorm}; they further utilize the conclusions to perform forward reasoning \cite{sole} or to enhance the KGE \cite{itere}. However, in most KGs, the facts are deterministic. Therefore, we believe that rules are uncertain but conclusions are deterministic in knowledge reasoning, as shown in Fig. \ref{fig1}; each fact is only absolutely true or false. Previous methods associate each conclusion with a weight derived from the corresponding rule. In contrast, we propose inferring that all conclusions are true ($\emph{e.g.}$, $(Mary, nationality, USA)$) or not ($\emph{e.g.}$, $(Mike, sister\_of, Nancy)$) (the other fact, $\emph{i.e.}$, $(Mike, gender, male)$, indicates that Mike is not Nancy's sister) by jointly modeling the deterministic KG and soft rules.

Specifically, we first mine soft rules from the knowledge graph and then infer conclusions as candidate facts. Second, the KG, conclusions, and weighted rules are also used as resources to learn embeddings. Third, through the definition of deterministic conclusion loss, the conclusion labels are modeled as 0-1 variables, and the confidence loss of a rule is also used to constrain the conclusions. Finally, the embedding learning stage removes the noise in the candidate conclusions, and then the proper conclusions are added back to the original KG. The above steps are performed iteratively. We empirically evaluate the proposed method on public datasets from two real large-scale KGs: DBpedia and Freebase. The experimental results show that our method Iterlogic-E (\textbf{Iter}ative using \textbf{logic} rule for reasoning and learning \textbf{E}mbedding) achieves state-of-the-art results on multiple evaluation metrics. Iterlogic-E also achieves improvements of  2.7\%/4.3\% in mean reciprocal rank (MRR) and 2.0\%/4.0\% in HITS@1 compared to the state-of-the-art model.

In summary, our main contributions are as follows:

\begin{itemize}

\item We propose a novel KGC method, Iterlogic-E, which jointly models logical rules and KGs in the framework of a KGE. Iterlogic-E combines the advantages of both rules and embeddings in knowledge reasoning. Iterlogic-E models the conclusion labels as 0-1 variables and uses a confidence regularizer to eliminate the uncertain conclusions.
\item We propose a novel iterative learning paradigm that achieves a good balance between efficiency and scalability. Iterlogic-E not only makes the KG denser but can also filter incorrect conclusions.
\item Compared with traditional reasoning methods, Iterlogic-E is more interpretable in determining conclusions. It not only knows why the conclusion holds but also knows which is true and which is false.
\item We empirically evaluate Iterlogic-E with the task of link prediction on multiple benchmark datasets. The experimental results indicate that Iterlogic-E can achieve state-of-the-art results on multiple evaluation metrics. The qualitative analysis proves that Iterlogic-E is more robust for rules with different confidence levels.

\end{itemize}

\section{Related Work}
Knowledge reasoning aims to infer certain entities over KGs as the answers to a given query. A query in KGC is a head entity $h$ (or a tail entity $t$) and a relation $r$. Given $(h,r, ?)$ (or $(?, r,t)$), KGC aims to find the right tail entity $t$ (or head entity $h$) in the KG that satisfies the triple $(h,r,t)$. Next, we review the three most relevant classes of KGC methods.

\subsection{Rule-Based Reasoning}
Logical rules can encode human knowledge compactly, and early knowledge reasoning was primarily based on first-order logical rules. Existing rule-based reasoning methods have primarily utilized search-based inductive logic programming (ILP) methods, usually searching and pruning rules. Based on the partial completeness assumption, AMIE \cite{amie} introduces a revised confidence metric, which is well suited for modeling KGs. By query rewriting and pruning, AMIE+ \cite{amie+} is optimized to expand to larger KGs. Additionally, AMIE+ improves the precision of the forecasts by using joint reasoning and type information. In this paper, we employ AMIE+\footnote{\url{https://github.com/lajus/amie}} to mine horn rules from a KG. Rule-based reasoning methods can be combined with multiple probability graph models. A Markov logic network (MLN) \cite{MLN} is a typical model. Based on preprovided rules, it builds a probabilistic graph model and then learns the weights of rules. However, due to the complicated graph structure among triples, the reasoning in an MLN is time-consuming and difficult, and the incompleteness of KGs also impacts the inference results. In contrast, Iterlogic-E uses rules to enhance KGEs with more effective inference.

\subsection{Embedding-Based Reasoning}
Recently, embedding-based methods have attracted much attention; they aim to learn distributed embeddings for entities and relations in KGs. Generally, current KGE methods can be divided into three classes: 1) translation-based models that learn embeddings by translating one entity into another entity through a specific relation \cite{transe,transms}; 2) compositional models that use simple mathematical operations to model facts, including linear mapping \cite{rotate}, bilinear mapping \cite{dismult, complex, analogy}, and circular correlation \cite{hole}; 3) neural network-based models that utilize a multilayer neural structure to learn embeddings and estimate the plausibility of triples with nonlinear features, for example, R-GCN \cite{r-gcn}, ConvE \cite{conve} and and so on \cite{soon1,soon2,soon3}. The above methods learn representations based only on the triples existing in KGs, and the sparsity of data limits them. To solve this problem and learn semantic-rich representations, recent works further attempted to incorporate information beyond triples, \emph{e.g.}, contextual information \cite{context}, entity type information \cite{entitytype1,entitytype2}, ontological information \cite{ontology}, taxonomic information \cite{taxonomic}, textual descriptions \cite{textual} and hierarchical information \cite{HAKE}. In contrast, the proposed Iterlogic-E uses embeddings to remove incorrect conclusions obtained by rules, which combines the advantages of rules and embeddings.

\subsection{Hybrid Reasoning}
Both rule-based and embedding-based methods have advantages and disadvantages. Recent works have integrated these two kinds of reasoning methods. Guo et al. \cite{KALE} attempted to learn a KGE from rule groundings and triples together. Wang et al. \cite{TARE} used asymmetric and transitive information to approximately order relations by maximizing the margin between negative and positive logical rules. Zhang et al. \cite{sole} and Guo et al. \cite{RUGE} obtained KGEs with supervision from soft rules, proving the effectiveness of logical rules. Qu et al. \cite{plogicnet} used an MLN to model logical rules and inferred new triples to enhance KGEs. Guo et al. \cite{slre} enhanced KGEs by injecting grounding rules. Niu et al. \cite{CAKE} enhanced KGEs by extracting commonsense from factual triples with entity concepts. In addition, some previous methods that enhance embeddings by iterative learning were studied in early works. Zhang et al. \cite{itere} aimed to improve a sparse entity representation through iterative learning and update the confidence of rules through embeddings. In contrast, Iterlogic-E models the conclusion labels as 0-1 variables and uses confidence regularization loss to eliminate the uncertain conclusions. Such labels are easier to train on.

\section{The Proposed Method}

\setcounter{figure}{1}
 \begin{figure*}[ht]
 \centering
 \includegraphics[width=1.0\textwidth]{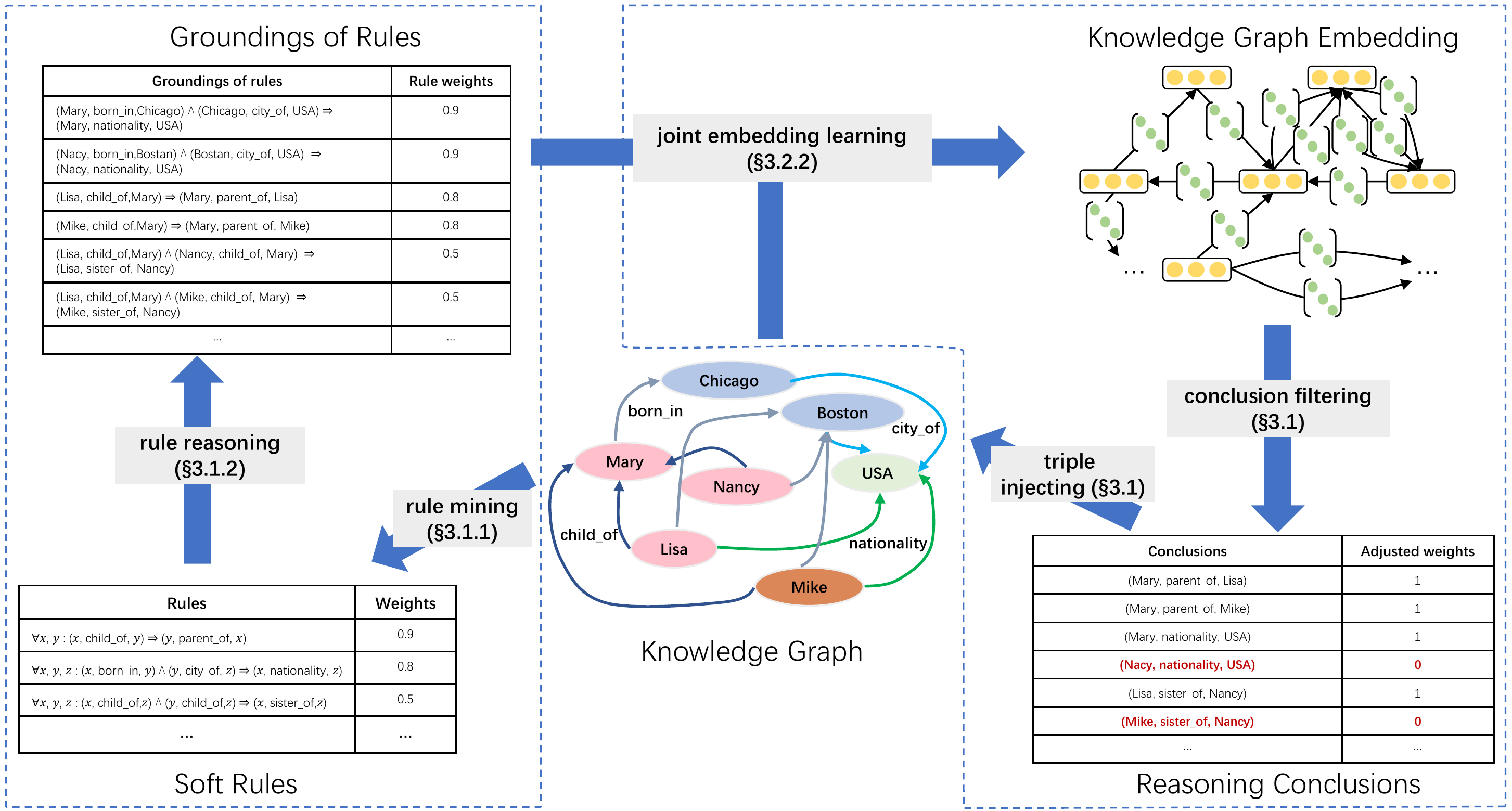}\\
  \caption{The framework details Iterlogic-E with two iterative stages: (i) rule mining and reasoning and (ii) embedding learning. Stage (i) generates rules and grounding rules to obtain new conclusions. Stage (ii) jointly models the conclusions of grounding rules and the KG in learning embeddings. After embedding learning, the conclusions are injected into the KG, and then the rule reasoning module is executed to start the next round of iterative training}
  \label{fig2}
\end{figure*}

This section introduces our proposed method Iterlogic-E. We first give an overview of our method, including the entire iterative learning process. Then, we detail the two parts of Iterlogic-E: rule mining and reasoning and embedding learning. Last, we discuss the space and time complexity of Iterlogic-E and discuss connections to related works \cite{itere, sole}.

\subsection{Overview}
Given a KG $\mathcal{G} =\{\mathcal{E}, \mathcal{R}, \mathcal{T}\}$, $\mathcal{T}=\{(h,r,t)\}$, $r \in \mathcal{R}$ is a relation and $h,t \in \mathcal{E}$ are entities. As discussed in Section~\ref{sec1}, on the one hand, embedding learning methods do not make full use of logical rules and suffer from data sparsity. On the other hand, precise rules are difficult to obtain efficiently and cannot cover all facts in KGs. Our goal is to improve the embedding quality by explicitly modeling the reasoned conclusions of logical rules, removing incorrect conclusions, and improving the confidence of the rules at the same time. Figure \ref{fig2} shows the overview of Iterlogic-E given a toy knowledge graph. Iterlogic-E is a general framework that can fuse different KGE models.

Iterlogic-E has two iterative steps: (i) rule mining and reasoning and (ii) embedding learning. In the rule mining and reasoning step, there are two modules: rule mining and rule reasoning. The mining configuration, such as the maximum length of rules and the confidence threshold of rules, and KG triples are input to the rule mining module. Then, it automatically obtains the soft rules from these inputs. The KG triples and the extracted soft rules are input into the rule reasoning module to infer new triples. After that, the new triples are appended to the embedding learning step as candidate conclusions. In the embedding learning step, relations are modeled as a linear mapping operation, and triple plausibility is represented as the correlation between the head and tail entities after the operation. Finally, the incorrect conclusions are filtered out by labeling the conclusions with their scores\footnote{In the experiments, we choose the conclusion with a normalized score of more than 0.99 as the true conclusion.}. The right conclusion will be added back to the original KG triples, and then the rule reasoning module is performed to start the next cycle of iterative training.

\subsection{Rule Mining and Reasoning}
The first step is composed of the rule mining module and the rule reasoning module. We introduce these two modules in detail below.

\subsubsection{Rule Mining}
We extract soft rules from the KG using the state-of-the-art rule mining method AMIE+ \cite{amie+} in this module. AMIE+ applies principal component analysis (PCA) confidence to estimate the reliability of a rule since its partial completeness assumption is more suited to real-world KGs. Additionally, AMIE+ defines a variety of restriction types to help extract applicable rules, \emph{e.g.}, the maximum length of the rule. After the rule mining module receives the KG triples and the mining configuration, it executes the AMIE+ algorithm and outputs soft rules. Although rules can be re-mined in each iteration, we only run the rule mining module once for efficiency reasons.

\subsubsection{Rule Reasoning}
\label{Rule Reasoning}
The logical rule set is denoted as $\mathcal{F} =\{(f, c)\}$. $f$ is in the form of $\forall x,y,z: (x,r_{p1},y)\wedge (y, r_{p2}, z) \overset{c}{\Rightarrow}(x, r_c, z)$. $x$, $y$ and $z$ represent variables of different entities, and $r_{p1}$, $r_{p2}$ and $r_{r_c}$ represent different relations. The left side of the symbol $\Rightarrow$ is the premise of the rule, which is composed of several connected atoms. The right side is only a single atom, which is the rule's conclusion. The horn rules are closed \cite{dismult}, where continuous relations share the intermediate entity and the first and last entities of the premise appear as the head and tail entities of the conclusion. Such rules can provide interpretive insights. A rule's length is equal to the number of atoms in the premise. For example, $\forall x,y,z: (x, born_in, y)\wedge (y, city, z) \overset{0.8 }{\Rightarrow} (x, nationality, z)$ is a length-2 rule. This rule reflects the reality that, most likely, a person's nationality is the country in which he or she was born. The rule $f$ has a confidence level of 0.8. The higher the confidence of the rule, the more likely it is to hold.

The reasoning procedure consists of instantiating the rule's premise and obtaining a large number of fresh conclusion triples. One of the most common approaches is forward chaining, also known as the match-select-act cycle, which works in three-phase cycles. Forward chaining matches the currently existing facts in the KG with all known rule premises in one cycle to determine the rules that can be satisfied. Finally, the selected rule's conclusions are derived, and
if the conclusions are not already in the KG, they are added as new facts.
 This cycle should be repeated until no new conclusions emerge. However, if soft rules are used, forward chain reasoning will lead to incorrect conclusions. Therefore, we run
one reasoning cycle in every iteration.

\subsection{Embedding Learning}
In this section, we present a joint embedding learning approach that allows the embedding model to learn from KG triples, conclusion triples, and soft rule confidence all at the same time. First, we will examine a basic KGE model, and then we will describe how to incorporate soft rule conclusions. Finally, we detail the overall training goal.

\subsubsection{A Basic KGE Model}
Different KGE models have different score functions that aim to obtain a suitable function to map the triple score to a continuous true value in [0, 1], \emph{i.e.}, $\phi:\mathcal{E}\times\mathcal{R}\times\mathcal{E}\rightarrow(0,1)$, which indicates the probability that the triple holds. We follow \cite{sole,slre} and choose ComplEx \cite{complex} as a basic KGE model. It is important to note that our proposed framework can be combined with an arbitrary KGE model. Theoretically, using a better base model can continue improving performance. Therefore, We also experiment with RotatE as a base model. Below we take ComplEx as an example to introduce. ComplEx assumes that the entity and relation embeddings exist in a complex space, \emph{i.e.}, $e\in\mathbb{C}^d$ and $r\in\mathbb{C}^d$, where $d$ is the dimensionality of the complex space. Using plurals to represent entities and relations can better model antisymmetric and symmetric relations (\emph{e.g.}, kinship and marriage) \cite{complex}. Through a multilinear dot product, ComplEx scores every triple:
\begin{equation}
F(h,r,t) = {\rm Re}(h^\mathrm{T}{\rm diag}(r)\bar{t}) = {\rm Re}({\textstyle \sum_i}[h]_i[r]_i[t]_i),
\label{e2}
\end{equation}

where the ${\rm Re(\cdot)}$ function takes the real part of a complex value and the ${\rm diag(\cdot)}$ function constructs a diagonal matrix from $r$; $\bar{t}$ is the conjugate of t; and $[\cdot]_i$ is the $i$-th entry of a vector.
To predict the probability, ComplEx further uses the sigmoid function for normalization:
\begin{equation}
\phi(h,r,t)=\sigma(F(h,r,t))=\sigma({\rm Re}(h^\mathrm{T}{\rm diag}(r)\bar{t})),
\end{equation}
where $\sigma(\cdot)$ is the sigmoid function. By minimizing the logistic loss function, ComplEx learns the relation and entity embeddings:

\begin{equation}
 \sum_{(h,r,t)\in \mathcal{T}\cup\mathcal{T}^\prime}\rm{log}(1+{\rm exp}(-y_{hrt}\cdot f(h,r,t))),
\end{equation}
where $\mathcal{T}^\prime$ is a set of sampled negative examples and $y_{hrt}$ is the label of a positive or negative triple.

\subsubsection{Joint Modeling KG and Conclusions of Soft Rules}
To model the conclusion label as a 0-1 variable, based on the current KGE model's scoring function, we follow ComplEx and use the function $f(\cdot)$ as the scoring function for conclusion triples:
\begin{equation}
S_i = \sigma(F(h_i,r_i,t_i)), (h_i,r_i,t_i)\in \mathcal{C}_f,
\end{equation}
where $\mathcal{C}_f$ is the set of conclusion triples derived from rule $f$ and $F(\cdot)$ is the score function defined in Equation \eqref{e2}. Aiming to regularize this scoring function so that it approaches 0 or 1, and to distinguish between true and false conclusions, we use a quadratic function with a symmetry axis of 0.5. Therefore, the conclusion score is the smallest when it is close to 0 or 1. Therefore, we define the deterministic conclusion loss $L_{dc}$ as follows:
\begin{equation}
L_{dc}=-\frac{1}{\left |\mathcal{C}_f\right |}\sum_{(h_i,r_i,t_i)\in \mathcal{C}_f} \left \| S_i-0.5 \right \|^2.
\end{equation}

According to the definition of rule confidence in \cite{amie}, the confidence of a rule $f$ in a KB $\mathcal{G}$ is the proportion of true conclusions among the true conclusions and false conclusions. Therefore, we can define the confidence loss of a rule as follows:

\begin{equation}
L_{rc} = \left \|\frac{1}{\left |\mathcal{C}_f\right |}\sum_{(h_i,r_i,t_i)\in \mathcal{C}_f}  S_i-c_f \right \|^2,
\end{equation} where $c_f$ is the confidence of rule $f$. Therefore, the loss of the conclusions of all the rules $L_{ac}$ can be defined as follows:

\begin{equation}
L_{ac} = \frac{1}{\left |\mathcal{F}\right |}\sum_{f\in \mathcal{F}}(L_{dc} + L_{rc}),
\end{equation}
where $\mathcal{F}$ is the set of all rules. To learn the KGE and rule conclusions at the same time, we minimize the global loss over a soft rule set $\mathcal{F}$ and a labeled triple set $\mathcal{L}={(x_l,y_l)}$ (including negative and positive examples). The overall training objective of Iterlogic-E is:

\begin{align}
\label{eq7}
\min_\theta\frac{1}{\left | \mathcal{L} \right |} & \sum_{(x_l,y_l)\in \mathcal{L}}L(-f(x_l)\cdot y_l) \notag\\
& +
\frac{1}{\left |\mathcal{F}\right |}\sum_{f\in \mathcal{F}}
(-\frac{1}{\left |\mathcal{C}_f\right |}\sum_{(h_i,r_i,t_i)\in \mathcal{C}_f} \left \| S_i-0.5 \right \|^2 \notag\\
& + \left \|\frac{1}{\left |\mathcal{C}_f\right |}\sum_{(h_i,r_i,t_i)\in \mathcal{C}_f}  S_i-c_f \right \|^2) .
\end{align}
where the $f(\cdot)$ function denotes the score function and $L(x) = \rm{log}(1 + \rm{exp}(x))$ is the soft-plus function. In Algorithm \ref{alg1}, we detail the embedding learning procedure of our method. To avoid overfitting, we further impose $l_2$ regularization on embedding $\Theta$. Following \cite{complex-nne,slre}, we also imposed nonnegative constraints (NNE) on the entity embedding to learn more effective features.


\begin{algorithm}[H]
\caption{Iterative learning algorithm of Iterlogic-E}
\SetAlgoLined
\LinesNumbered 
\label{alg1}
\KwIn{
KG triples $\mathcal{T}=\{(e_i,r_k,e_j)\}$, logical rules $\mathcal{F}=\{(f_p,c_p)\}$, the number of iterative
learning steps
$M$.

}
\KwOut{Relation and entity embeddings $\Theta$}
Randomly initialize relation and entity embeddings $\Theta(0)$\;
 \For{$n\leftarrow 1$ to $N$ }{
  $\mathcal{C}\leftarrow \varnothing$\;
  \If{n/[N/M] == 0}
  {Generate a set of conclusions $\mathcal{C}^\prime=\{(e_i^\prime,r_k^\prime,e_j^\prime)\}$ by rule grounding from $\mathcal{T}, \mathcal{F}$\;
  $\mathcal{C}=\mathcal{C}\cup \mathcal{C}^\prime$\;
  }
  Sample a mini-batch $\mathcal{T}^b, \mathcal{C}^b$ from $\mathcal{T}, \mathcal{C}$\;
  Generate a set of negative triples $\mathcal{T}^b_{neg}$\;
  $\mathcal{L}^b\leftarrow \varnothing$\;
  \For{each $x_l\in \mathcal{T}^b_{neg} \cup \mathcal{T}^b \leftarrow 1$ to $N$ }{
  $y_l = +1/-1 $\;
  $\mathcal{L}^b\leftarrow \mathcal{L}^b \cup {(x_l,y_l)}$\;
  }
  $\Theta^{(n)}\leftarrow\Theta^{(n-1)}-\eta(\frac{1}{\left |\mathcal{L}^b  \right |} \sum_{(x_l,y_l)\in \mathcal{L}^b}\bigtriangledown_\Theta L(-f(x_l)\cdot y_l)
  +
\frac{1}{\left |\mathcal{F}\right |}\sum_{f\in \mathcal{F}}
(-\frac{1}{\left |\mathcal{C}_f\right |}\sum_{(h_i,r_i,t_i)\in \mathcal{C}_f} \bigtriangledown_\Theta\left \| S_i-0.5 \right \|^2
+ \bigtriangledown_\Theta\left \|\frac{1}{\left |\mathcal{C}_f\right |}\sum_{(h_i,r_i,t_i)\in \mathcal{C}_f}  S_i-c_f \right \|^2)$ \commentSt*[r]{\emph{cf.} Eq. \eqref{eq7}}
  $\mathcal{C}_t\leftarrow \varnothing$\;
  \For{each $x_m\in\mathcal{C}$}{
  \If{$\sigma(f(x_m))>=0.99$}{$\mathcal{C}_t\cup(x_m)$\;}
  }
  $\mathcal{T} = \mathcal{T}\cup\mathcal{C}_t$\;
 }
\textbf{return} $\Theta^{(N)}$
\end{algorithm}

\subsection{Discussion}
\subsubsection{Complexity}
 In the embedding learning step, we represent relations and entities as complex value vectors, following ComplEx.
As a result, the space complexity is $O(n_ed+n_rd)$, where $d$ is the embedding space's dimensionality. The number of relations is $n_r$, and the number of entities is $n_e$. Each iteration of the learning process has a time complexity of $O(n_ld+n_cd)$, where $n_l/n_c$ is the number of new conclusions or the number of labeled triples in a mini-batch, as shown in Algorithm \ref{alg1}. Iterlogic-E is similar to ComplEx in that its space and time complexity increase linearly with $d$.
The number of new conclusions in a minibatch is usually considerably lower than the number of initial triples; \emph{i.e.}, $n_c\ll n_l$. As a result, Iterlogic-E's time complexity is very close to that of ComplEx, which needs only $O(n_ld)$ per iteration. Because of the rule mining module's great efficiency and practical constraints, such as the PCA confidence threshold not being lower than 0.5 and the length of rules not exceeding two, the rule grounding stage's space and time complexity is trivial compared to that of the embedding learning stage. Therefore, we may disregard it when considering the space and time complexity of Iterlogic-E.

\subsubsection{Connection to Related Works}

IterE \cite{itere} also uses iterative learning, which defines several types of rules with OWL2, but IterE does not change the process of embedding learning and is limited by rules that will yield many noisy conclusions. IterE uses a pruning strategy that utilizes traversal and random selection to obtain rules. Moreover, they only improve the prediction effect of sparse entities but not well on standard datasets. By contrast, Iterlogic-E uses the SOTA rule mining system \cite{amie+} to mine high-confidence rules, and the quality of the rules obtained in this way is higher because it uses the KG to fully evaluate the reliability of the rules. SoLE \cite{sole} enhances KGE by jointly modeling the groundings of rules and facts and directly utilizes uncertain rules for forward chain reasoning without eliminating incorrect grounding. Moreover, SoLE uses t-norm based fuzzy logic \cite{tnorm} to model grounding, which will greatly increase the time complexity. The method we propose avoids the above mentioned problems without increasing the number of parameters. SLRE \cite{slre} uses rule-based regularization that merely enforces relation to satisfying constraints introduced by soft rules. However, it does not use rules for reasoning and can not benefit from the interpretability and accuracy advantages. Moreover, SLRE has strict requirements on the form of the rules, while our method can utilize various forms of rules more simply and flexibly via the rule reasoning module.

\section{Experiments and Results}
\subsection{Datasets}
Iterlogic-E is tested on two common datasets: FB15K and DB100K. The first is based on Freebase, which was released by Bordes et al. \cite{transe}. The second was taken from DBpedia by Ding et al. \cite{complex-nne}, and it includes 99,604 entities and 470 relations. For model training, hyperparameter tuning and evaluation, we utilize fixed training, validation, and test sets on both datasets.

With each training dataset, we obtain soft rules and examine rules with a length of no more than 2 to allow efficient extraction. These rules, together with their confidence levels, are automatically retrieved from each dataset's training set using AMIE+ \cite{amie+}, and only those with confidence levels greater than 0.8 are used. Shorter rules are thought to more directly represent logical connections among relations. Therefore, we remove longer rules when all of their relations also exist in shorter ones. Table \ref{tab1} summarizes the datasets' comprehensive statistics, and Table \ref{tab2} also includes several rule instances. We can observe from the statistics that the number of rules on both datasets is extremely minimal when compared to the number of triples.

\begin{table}[H]
  \caption{Statistics of the datasets, where the columns represent the numbers of entities, relations, training/validation/test triples, and soft rules}
 \setlength{\tabcolsep}{1mm}
  \label{tab1}
  \begin{tabular}{rrrrr}
    \toprule
    Dataset& \# Ent& \# Rel& \# Train/Valid/Test & \# Rule\\
    \midrule
    FB15K & 14,951&1,345&483,142/50,000/59,071&441\\
    DB100K & 99,604& 470&597,572/50,000/50,000&25\\

  \bottomrule
\end{tabular}
\end{table}

\begin{table}[H]
  \caption{Examples of rules, with confidences, that were extracted from FB15K (top) and DB100K (bottom)}
  \setlength{\tabcolsep}{1mm}
  \label{tab2}
  \begin{tabular}{l}
    \toprule
    /location/contains$(y,x)\overset{0.84}{\Rightarrow}$/location/containedby(x,y) \\
    /production\_company/films$(y,x) \overset{0.89}{\Rightarrow}$/location/\\
    containedby(x,y)/hud\_county\_place/place$(x,y)\wedge$ \\
    hud\_county\_place/county$(y,z)$ \\ $\overset{1.0}{\Rightarrow}$
    /hud\_county\_place/county$(x,z)$\\
    \midrule
    sisterNewspaper$(x,y)\wedge $sisterNewspaper$(z,y) \overset{0.82}{\Rightarrow}$\\
    sisterNewspaper$(x,z)$\\
    distributingCompany$(x,y)\overset{0.91}{\Rightarrow}$distributingLabel$(x,y)$ \\
    nationality$(x,y)\overset{0.99}{\Rightarrow}$stateOfOrigin$(x,y)$ \\
  \bottomrule
\end{tabular}
\end{table}

\begin{table*}[!htb]
\caption{Link prediction results on the test sets of FB15K and DB100K}
\label{tab3}
\begin{tabular*}{1.0\hsize}{@{}@{\extracolsep{\fill}}llcccccccc@{}}
\toprule
&\multirow{2}{*}{Method}&\multicolumn{4}{c}{FB15K}&\multicolumn{4}{c}{DB100K}\\
\cmidrule(r){3-6}\cmidrule(r){7-10}
&&MRR&HITS@1&HITS@3&HITS@10&MRR&HITS@1&HITS@3&HITS@10\\
\midrule
1&TransE \cite{transe}& 0.380&0.231&0.472&0.641&0.111&0.016&0.164&0.270\\
2&DistMult \cite{dismult}& 0.654&0.546&0.733&0.824&0.233&0.115&0.301&0.448\\
3&HolE  \cite{hole}& 0.524&0.402&0.613&0.739&0.260&0.182&0.309&0.4118\\
4&ComplEx  \cite{complex}& 0.627&0.550&0.671&0.766&0.272&0.218&0.303&0.362\\
5&ANALOGY  \cite{analogy}& 0.725&0.646&0.785&0.854&0.252&0.143&0.323& 0.427\\
6&ComplEx-NNE  \cite{complex-nne}& 0.727& 0.659& 0.772& 0.845& 0.298& 0.229& 0.330& 0.426\\
7&ComplEx-CAS  \cite{ComplEx-CAS}& -& - &-&0.866& -& - &-& -\\
8&RotatE    \cite{rotate}& 0.664&0.551&0.751&0.841& 0.327& 0.200 &0.417& 0.526\\
9&HAKE\cite{HAKE}& 0.408&0.312&0.463&0.579& - & - &-& -\\
10&CAKE\cite{CAKE}& 0.741&0.646&0.825&0.896& -& - &-& -\\
11&R-GCN+   \cite{r-gcn}& 0.696&0.601&0.760&0.842&-&-& -& -\\
12&ConvE  \cite{conve}& 0.745& 0.670 &0.801 &0.873 &- &-\\
13&DPMPN\cite{DPMPN} & 0.764&0.726 &0.784 &0.834 &- &- &- &-\\
\midrule
14&BLP  \cite{BLP} & 0.242&0.151&0.269&0.424&-&-& -& -\\
15&MLN   \cite{MLN}& 0.321&0.210&0.370&0.550&-&-& -& -\\
\midrule
16&PTransE  \cite{ptranse}& 0.679& 0.565& 0.768& 0.855& 0.195& 0.063& 0.278 &0.416\\
17&KALE  \cite{KALE}& 0.518& 0.382& 0.606& 0.756 &0.249& 0.100& 0.346& 0.497\\
18&ComplEx$^R$  \cite{complex-r} & - &-& -& -& 0.253& 0.167& 0.294& 0.420\\
19&TARE   \cite{TARE} & 0.781& 0.617& 0.728& 0.842& - &- &- &-\\
20&RUGE  \cite{RUGE} & 0.768& 0.703& 0.815& 0.865 &0.246 &0.129& 0.325 &0.433\\
21&ComplEx-NNE+AER \cite{complex-nne} & 0.801 &0.757& 0.829 &0.873& 0.311& 0.249 &0.339 &0.426\\
22&IterE  \cite{itere} & 0.576 &0.443 &0.665 &0.818& 0.274 &0.215 &0.299 &0.386\\
23&pLogicNet  \cite{plogicnet}&0.776& 0.706 &0.817& 0.885& - &- &- &-\\
24&SoLE  \cite{sole} &0.801 &0.764& 0.821 &0.867& 0.306& 0.248& 0.328 &0.418\\
25&SLRE  \cite{slre} &0.810& 0.774& 0.829 &0.871& 0.340& 0.261& 0.372&0.490\\
\midrule
&Iterlogic-E(ComplEx)&  0.814&0.778&0.835&0.873&0.374&\textbf{0.301}&0.409&0.509\\
&Iterlogic-E(RotatE)&  \textbf{0.837}&\textbf{0.794}&\textbf{0.868}&\textbf{0.904}&\textbf{0.387}&0.287&\textbf{0.449}&\textbf{0.559}\\
\bottomrule
\end{tabular*}
\end{table*}

\subsection{Link Prediction}
Our method was evaluated on link prediction. The goal of this task was to restore a missing triple $(e_i,r_k,?)$ with the tail entity $e_j$ or $(?,r_k,e_j)$ with the head entity $e_i$.

\subsubsection{Evaluation Protocol}
The standard protocol established by \cite{transe} is used for evaluation. The head entity $e_i$ is replaced with each entity for every test triple $(e_i,r_k,e_j)$, and the corrupted triple's score is calculated. We record the rank of the right entity $e_i$ by ranking these scores in decreasing order. The mean reciprocal rank (MRR) and the percentage of ranks no greater than N (H@N, N = 1, 3, 10) are used to evaluate the ranking quality of all test triples.

\begin{table*}[!htb]
\centering
\caption{Link prediction results on the test sets of FB15K-sparse}
\label{tab4}
\begin{tabular}{lcccc}
\toprule
\multirow{2}{*}{Method}&\multicolumn{4}{c}{FB15K-sparse}\\
\cmidrule(r){2-5}
&MRR&HITS@1&HITS@3&HITS@10\\
\midrule
TransE& 0.398& 0.258&0.486&0.645\\
DistMult& 0.600& 0.618&0.651&0.759\\
ComplEx& 0.616& 0.540&0.657&0.761\\
IterE& 0.628& 0.551&0.673&0.771\\
SoLE& 0.668 & 0.604 & 0.699 & 0.794 \\
\midrule
Iterlogic-E(ComplEx) & \textbf{0.674} & \textbf{0.611} & \textbf{0.701} & \textbf{0.800}\\
\bottomrule
\end{tabular}
\end{table*}

\subsubsection{Comparison Settings}
We compare the performance of our method to that of a number of previous KGE models, as shown in Table \ref{tab3}. The translation-based model (row 1), the compositional models utilizing basic mapping operations (rows 2-10) and the neural network-based models (rows 11-13) are among the first batch of baselines that rely only on triples seen in the KGs.
The second batch of baselines are rule-based methods (rows 14-15). The last batch of baselines further incorporates logical rules (rows 16-25).

\subsubsection{Implementation Details}
On FB15K and DB100K, we compared Iterlogic-E against all of the baselines. We immediately obtained the results of a set of baselines on FB15K and DB100K from SoLR, SLRE and CAKE. We reimplemented ComplEx on the PyTorch framework based on thebased on the code code\footnote{\url{https://github.com/DeepGraphLearning/KnowledgeGraphEmbedding}} supplied by \cite{rotate} since our approach was dependent on it. Then, depending on our implementation, we provided the ComplEx result. Furthermore, the IterE result was tested on the sparse version of FB15K (FB15K-sparse) released by \cite{itere}, which included only sparse entities with 18,544 and 22,013 triples in the validation and test sets. Therefore, we reimplemented IterE on FB15K and DB100K based on the code and hyperparameters\footnote{\url{https://github.com/wencolani/IterE}} released by the author. As a result, we compared our approach to IterE and SoLE on the FB15K-sparse dataset. Both approaches use a logistic loss and optimize in the same way (SGD with AdaGrad). The other results of the baselines were obtained directly from prior literature. We tuned the embedding dimensionality $d$ within \{100, 150, 200, 250, 300\}, the number of negatives per positive triple $\eta$ within \{2, 4, 6, 8, 10\}, the initial learning rate $\gamma$ within $\{10^{-4}, 10^{-3}, 10^{-2}, 5\times10^{-2}, 10^{-1}\}$, and the L2 regularization coefficient $\mu$ within $\{10^{-5},3\times10^{-5},10^{-4},10^{-3}, 3\time10^{-3}, 10^{-2}\}$. We further tuned the margin within \{0.1, 0.2, 0.5, 1, 2, 5, 12, 18, 24\} for the approaches that utilize the margin-based ranking loss. The best hyperparameters were selected to maximize the MRR on the validation set. The best settings for Iterlogic-E were $d=300$, $\gamma=10^{-3}$, $\eta=10$, and $\mu=3\times10^{-5}$ on FB15K and $d=300$, $\gamma=10^{-4}$, $\eta=10$, and $\mu=10^{-4}$ on DB100K.

\subsubsection{Main Results}
The results of all compared methods on the test sets of FB15K, DB100K, and FB15K-sparse are shown in Tables \ref{tab3} and \ref{tab4}. For each test triple, the mean reciprocal rank or H@N value with N = 1, 3, 10 is utilized as paired data. The experimental results show that 1) Iterlogic-E outperforms numerous strong baselines in the vast majority of cases. This shows that Iterlogic-E can achieve very good accuracy. 2) Iterlogic-E significantly outperforms the basic models that use triples alone, and the improvement comes from the ability to learn the conclusions obtained by soft rules. 3) Iterlogic-E also beats many baselines that incorporate logical rules. Specifically, Iterlogic-E performs better than SoLE and IterE under most metrics. This demonstrates the superiority of Iterlogic-E in reducing the noise of candidate conclusions. 4) IterE can only enhance sparse entities, so the experimental results are much lower than those of other baseline models. However, Iterlogic-E is also effective on FB15K-sparse. 5) On DB100K, the improvements over SLRE and SoLE are more significant than those on FB15K. The reason for this is probably that the groundings of the rules on DB100K contain more incorrect conclusions. Simple rules between a pair of relations are adequate to capture these simple patterns on the FB15K dataset. 6) The performance of Iterlogic-E(ComplEx) is worse than some baselines in HIT@10, and we consider that this limitation is mainly due to the shortcomings of the base model ComplEx \cite{complex}. Sun et al. \cite{rotate} point out that ComplEx can not model the composition relation. We have experimented with replacing the base model with the RotatE model(Iterlogic-E(RotatE)), which is capable of modeling four relation patterns. The experimental results have been further improved, and our method consistently achieves the best results in all evaluation metrics.

\subsubsection{Ablation Study}
\begin{table}[H]
\centering
\caption{Ablation study}
\setlength{\tabcolsep}{0.5mm}
\label{tab5}
\begin{tabular}{lccccc}
\toprule
&\multirow{2}{*}{}&\multicolumn{4}{c}{DB100K}\\
\cmidrule(r){3-6}
&&MRR&HITS@1&HITS@3&HITS@10\\
\midrule
&Iterlogic-E& \textbf{0.374}& \textbf{0.301}&\textbf{0.409}&0.509\\
\midrule
1&w/o $l_2$ on $\Theta$& 0.323 & 0.278&0.342&0.409\\
2&w/o NNE & 0.371& 0.295&0.409&0.509\\
3&1+2& 0.324& 0.279&0.342&0.410\\
4&w/o IL & 0.372& 0.300&0.407&0.505\\
5&w/o IL+$L_{dc}$ & 0.347& 0.258&0.395&0.510\\
6&w/o IL+$L_{rc}$ & 0.351& 0.257&0.406&\textbf{0.515}\\
7&w/o IL+$L_{dc}$+$L_{rc}$ & 0.328& 0.279&0.348&0.422\\
8&Iterlogic-E*& 0.369& 0.298&0.404&0.502\\
9&ComplEx+AC+IL& 0.285& 0.223&0.312&0.407\\
10&ComplEx+WC+IL& 0.295& 0.231&0.327&0.422\\
\midrule
&ComplEx&0.272&0.218&0.303&0.362\\
\bottomrule
\end{tabular}
\end{table}

To explore the influence of different constraints and iterative learning, we perform an ablation study of Iterlogic-E on DB100K with 9 configurations in Table \ref{tab5}. The first and second variants, compared to the completed model Iterlogic-E, remove the non-negativity and $l_2$-norm constraints ($l_2$). The third setting is the combination setting. The fourth removes iterative learning (IL), which uses a rule to reason only once. The fifth, sixth, and seventh variants remove the additional loss item based on the sixth variant. The eighth setting is another variant of the Iterlogic-E model based on the ninth setting, which is that after the ComplEx fitting, according to the scores of the conclusions, the top $n$ (where $n$ is the rounded product of the rule and its confidence) conclusions of each rule are selected to be added to the KG to continue training. The ninth setting (AC) is to add all conclusions inferred from the rules as positive examples, and the tenth setting (WC) is to use the rule confidence as soft labels of conclusions.

 As seen in Table \ref{tab5}, we can conclude the following: 1) When removing NNE constraints, the performance of Iterlogic-E decreases slightly. Without $l_2$-norm constraints on entities, the performance of Iterlogic-E degrades by 2.3\% in H@1 and by 5.1\% in MRR.
 One explanation may be that $l_2$-norm constraints are sufficient to constrain embedding norms on DB100K. However, the performance will suffer dramatically if there are no $l_2$-norm constraints. 2) Removing iterative learning decreases performance slightly. One reason may be that the number of rules on DB100K is relatively small, so the number of conclusions added through iterative learning is relatively small. 3) Removing the additional loss item of the conclusions decreases performance slightly. This illustrates that Iterlogic-E can filter out incorrect conclusions and makes the KG dense. Surprisingly, even if we directly use ComplEx to filter and learn the conclusions that can achieve such high performance, this method is not as flexible as Iterlogic-E. 4) Compared with the basic model, all variants have different degrees of improvement. This demonstrates the critical importance of logical rules in link prediction tasks.

\subsection{Influence of the Number of Iterations}
\begin{figure}[H]
\centering
 \includegraphics[scale=0.5]{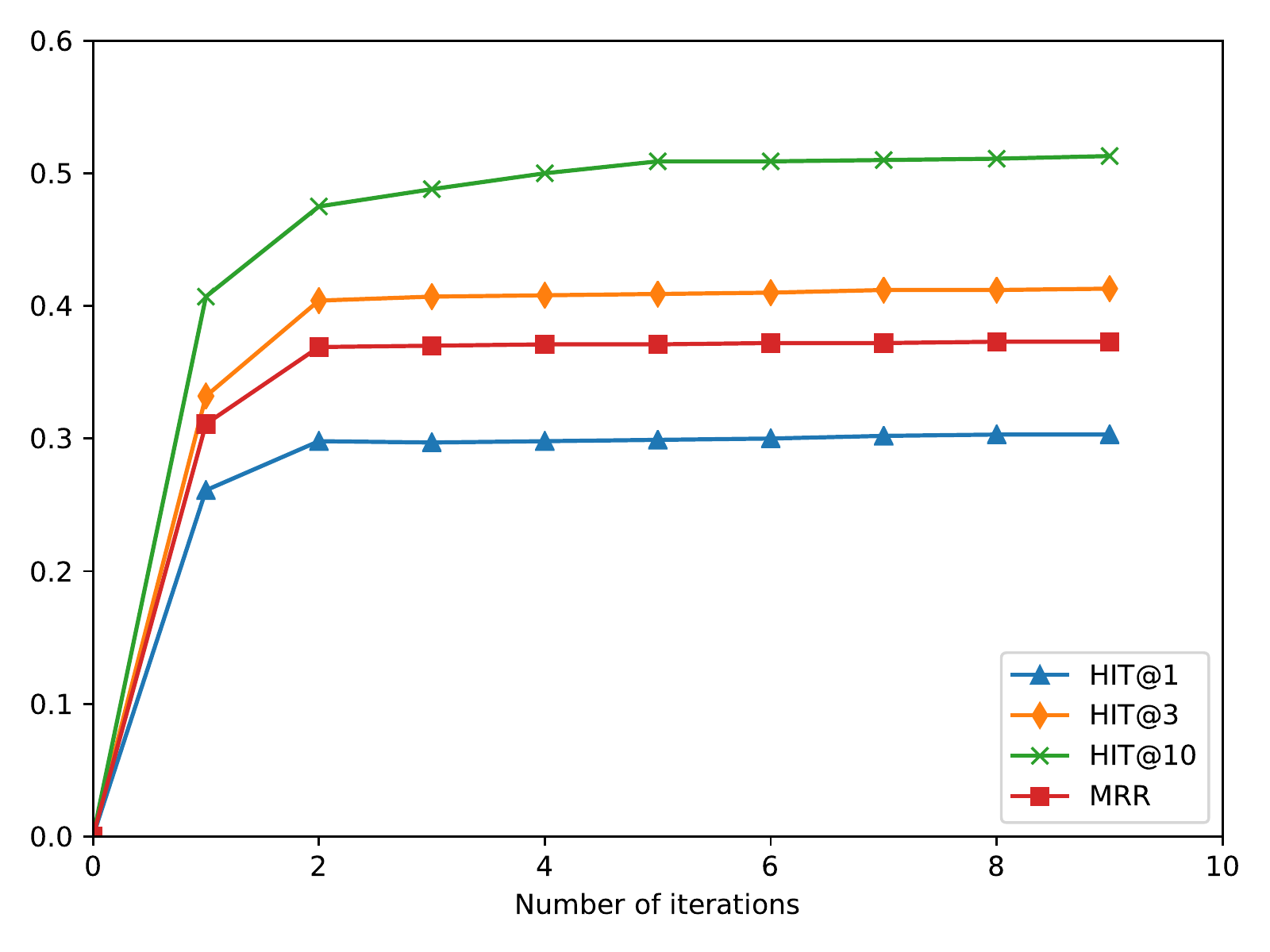}\\
  \caption{Link prediction results in different iterations}
  \label{fig3}
\end{figure}

To demonstrate how Iterlogic-E can enhance the embedding effect during the training process, we show the link prediction results on DB100K with different iterations. Figure \ref{fig3} shows that as the number of training iterations increases, the prediction results, including Hit@1, Hit@3, Hit@10, and MRR, will improve. From Fig. \ref{fig3}, we can infer the following: 1) Iterative learning enhances embedding learning since the quality of embeddings improved with time. 2) In the first two iterations, the embedding learning module was quickly fitted to the conclusions of the rules and the triples of the training set, and the prediction accuracy rapidly improved. 3) After two iterations, as the number of new conclusions decreased, the results of the inference tended to be stable, and the true conclusions and the initial KG triples  were well preserved in the embedding.

\subsection{Influence of Confidence Levels}

\begin{figure*}[!htb]
\centering
\subfigure[]{
\begin{minipage}[t]{0.4\linewidth}
\centering
\includegraphics[scale=0.4]{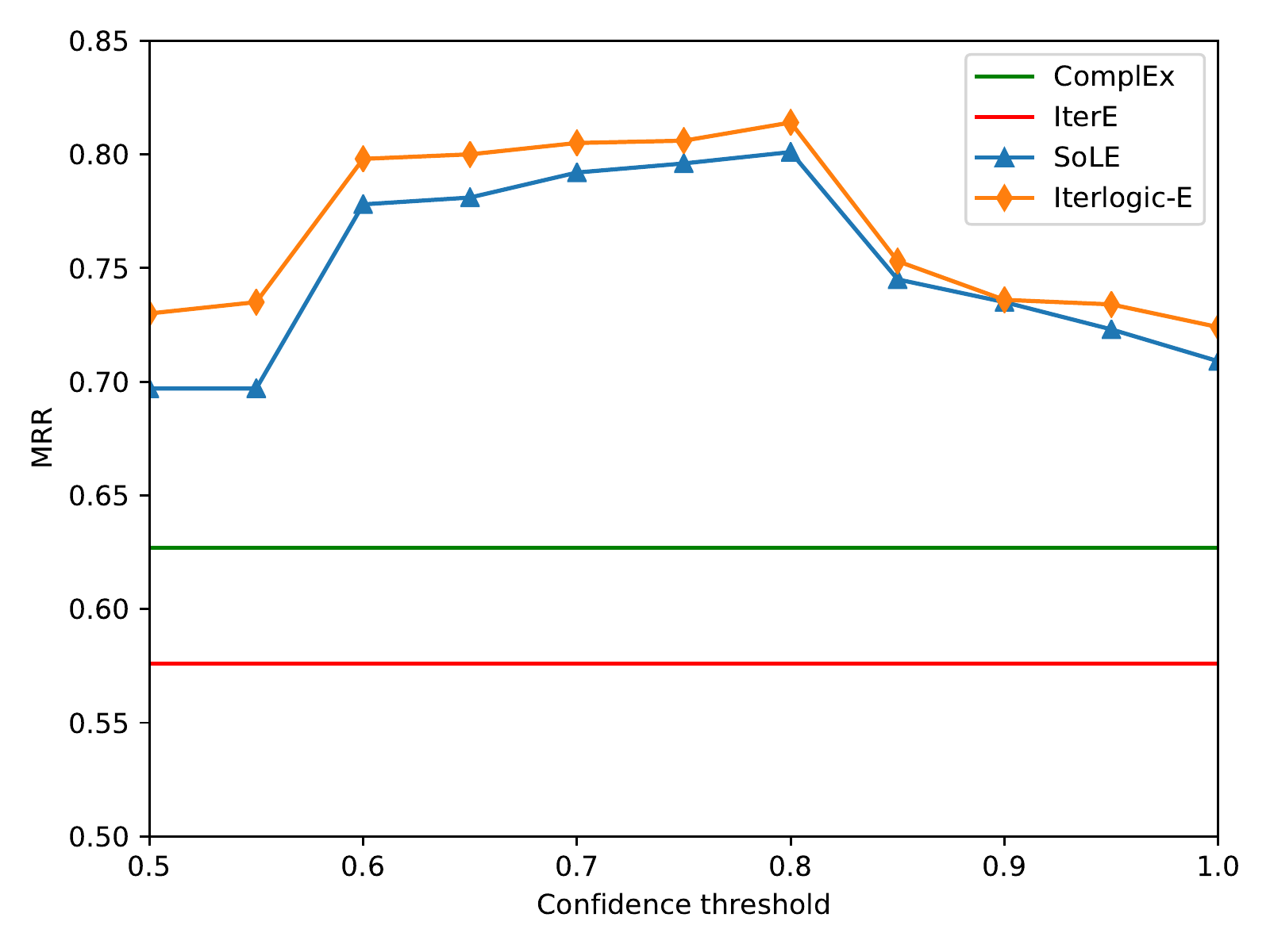}
\end{minipage}%
}%
\hspace{0.5in}
\subfigure[]{
\begin{minipage}[t]{0.4\linewidth}
\centering
\includegraphics[scale=0.4]{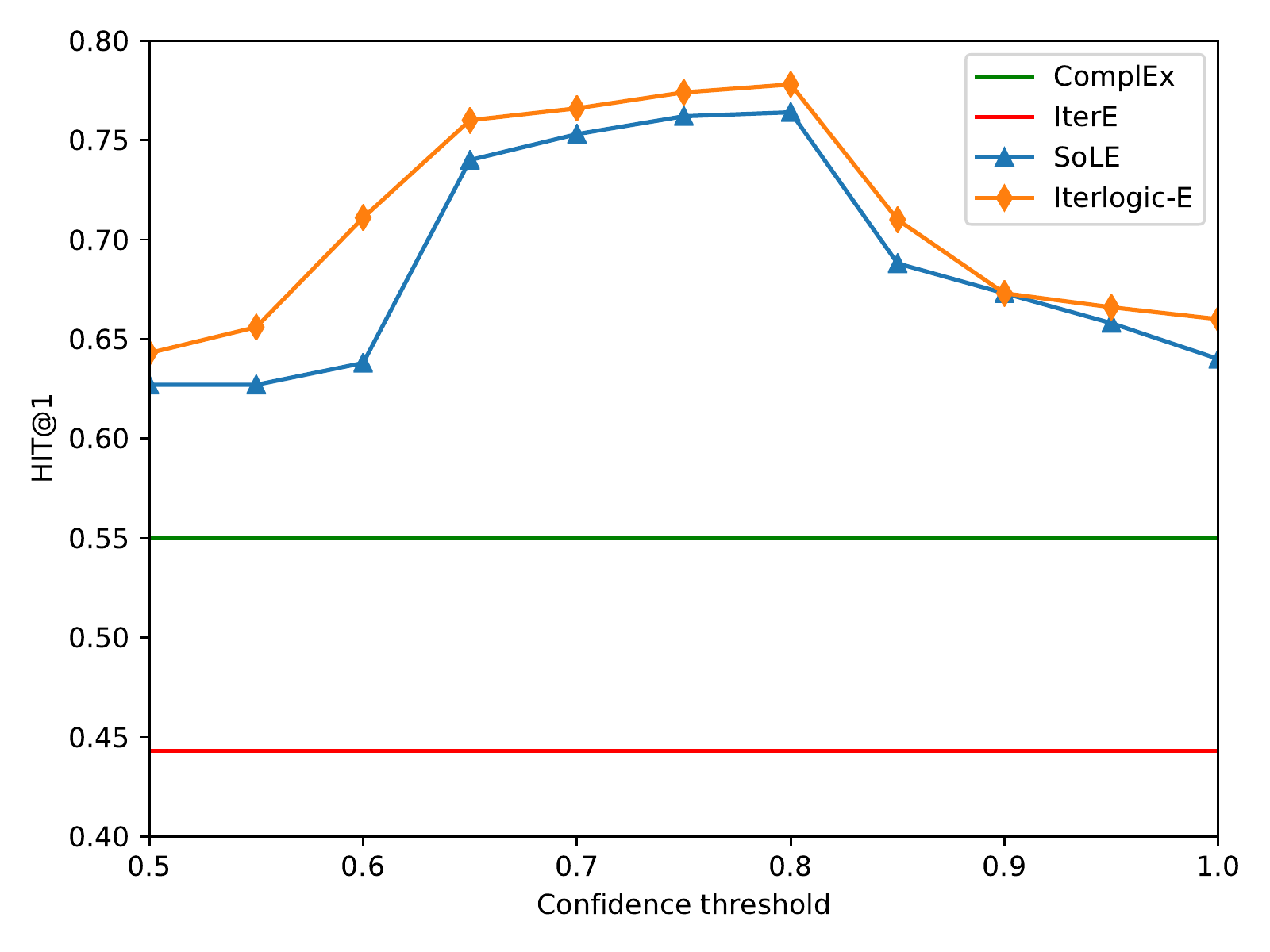}
\end{minipage}%
}%
\centering
\caption{
Results of MRR (\textbf{a}) and HITS@1 (\textbf{b}) achieved by Iterlogic-E with different confidence thresholds on FB15K}
\label{fig4}
\end{figure*}

Additionally, we evaluate the effect of the rules' confidence thresholds on FB15K. Since there are different rules and ComplEx does not merge rules, we refer only to their fixed results on FB15K. We set all hyperparameters to their optimum values and change the confidence threshold within [0.5, 1] in 0.05-step increments. Both SoLE and Iterlogic-E use rules with confidence levels greater than this threshold. Figure \ref{fig4} displays the MRR and H@1 values obtained by Iterlogic-E and other baselines on the FB15K test set. We make the following observations: 1) Iterlogic-E beats both ComplEx and IterE at varying confidence levels. This demonstrates that Iterlogic-E is sufficiently robust to deal with uncertain soft rules.

\subsection{Case Study}

\begin{table*}[!htp]

\caption{A case study with 4 conclusions (true or false as predicted by Iterlogic-E), which are reasoned on by 2 rules}
\label{tab6}
\begin{tabular}{p{3.5cm}|p{6cm}<{\centering}|p{6cm}<{\centering}}

\toprule
&True&False\\
\midrule
Conclusions&(Albany\_Devils,/hockey\_roster\_position/
position,Centerman)
&
(San\_Diego\_State\_Aztecs\_football,/hockey\_
roster\_position/position,Linebacker)\\

\midrule
Predicted by rules&\multicolumn{2}{c}{/sports\_team\_roster/position$(x,y)\overset{0.808}{\Rightarrow}$/hockey\_roster\_position/position$(x,y)$} \\
\midrule
Score change &5.4888 $\rightarrow$ 9.4039&-4.6518 $\rightarrow$ -8.3547\\
\midrule

Conclusions& (Chris\_Nurse,/football\_roster\_position/
team,Stevenage\_F.C.)
&
(Brett\_Favre,/football\_roster\_position/
team,Green\_Bay\_Packers)\\

\midrule
Predicted by rules&\multicolumn{2}{c}{/sports\_team\_roster/team$(x,y)\overset{0.847}{\Rightarrow}$/football\_roster\_position/team$(x,y)$}\\
\midrule
Score change &5.8123 $\rightarrow$ 9.5991&
-3.0131 $\rightarrow$ -11.3952\\

\bottomrule
\end{tabular}
\end{table*}

In Table \ref{tab6}, we present a case study with 4 conclusions (true or false as predicted by Iterlogic-E), which are inferred with 2 rules during training. Table \ref{tab6} shows some conclusions derived from rule inference and the score change (the average of the head entity prediction score and the tail entity prediction score) of the conclusions. Using the first conclusion as an example, the true conclusion is (Albany\_Devils,/hockey\_roster\_position/posit on, Centerman), which is obtained by the rule "/sports\_team\_roster/position$(x,y)\overset{0.808}{\Rightarrow}$/
hockey\_roster\_position/position$(x,y)$". The Albany Devils are a professional ice hockey team in the American Hockey League \footnote{\url{https://en.wikipedia.org/wiki/Albany_Devils}}, and the centerman is the center in ice hockey. Therefore, this is indeed a true conclusion. Compared to ComplEx, Iterlogic-E increased the score of this fact from 5.48 to 9.40. Also, (San\_Diego\_State\_Aztecs\_football,
/hockey\_roster\_position/position,Linebacker) can be inferred from the fact (San\_Diego\_State\_Aztecs\_football, /sports\_team\_roster/
team, Linebacker) by the same rule. However, the San Diego State Aztecs are a football team, not a hockey team, and this is an incorrect conclusion. Compared to ComplEx, Iterlogic-E decreased the score of this fact from -4.65 to -8.35. This illustrates that Iterlogic-E can distinguish whether the conclusion is true and can improve the prediction performance. Furthermore, Iterlogic-E has good interpretability, and we can understand why the conclusion is inferred by it.

\section{Conclusion and Future Work}
This paper proposes a novel framework that iteratively learns logical rules and embeddings, which models the conclusion labels as 0-1 variables. The proposed Iterlogic-E uses the confidences of rules and the context of the KG to eliminate the uncertainty of the conclusion
in the stage of learning embeddings.
Specifically, our method is based on iterative learning, which not only supplements conclusions but also filters incorrect conclusions, resulting in a good balance between efficiency and scalability. The evaluation on benchmark KGs demonstrates that the method can learn correct conclusions and improve against a variety of strong baselines. In the future, we would like to explore how to use embeddings to learn better rules and rule confidences than AMIE+. Additionally, we will continuously explore more advanced models to integrate rules and KGEs for knowledge reasoning.

\footnotesize
\itemsep=-3pt plus.2pt minus.2pt
\baselineskip=13pt plus.2pt minus.2pt


\label{last-page}
\end{multicols}
\label{last-page}
\end{document}